\documentclass[letterpaper]{article} 

\usepackage{fdiaz}

\usepackage{graphicx}

\newcommand{\metric}[0]{\mu}

\newcommand{\construct}[0]{\metric^*}

\newcommand{\metrics}[0]{\mathcal{M}}

\newcommand{\model}[0]{\pi}
\newcommand{\modelspace}[0]{\Pi}

\newcommand{\trainedmodel}[0]{\model(\trainset)}

\newcommand{\testset}[0]{\samplepeople}

\newcommand{\trainset}[0]{\mathcal{D}}
\newcommand{\trainseti}[1]{\trainset_{#1}}
\newcommand{\trainsetsizei}[1]{|\trainseti{#1}|}

\newcommand{\performancebase}[0]{\metric\left(\testset,\trainedmodel\right)}

\newcommand{\trueperformancebase}[0]{\construct\left(\people,\trainedmodel\right)}

\newcommand{\people}[0]{\mathcal{U}}
\newcommand{\samplepeople}[0]{U}

\newcommand{\samplingdist}[0]{\theta}
\newcommand{\support}[0]{\mathrm{supp}}

\usepackage{aaai24}  
\usepackage{times}  
\usepackage{helvet}  
\usepackage{courier}  
\usepackage[hyphens]{url}  
\usepackage{graphicx} 
\usepackage{natbib}  
\usepackage{caption} 
\usepackage{algorithm}
\usepackage{algorithmic}

\usepackage{newfloat}
\usepackage{listings}
\DeclareCaptionStyle{ruled}{labelfont=normalfont,labelsep=colon,strut=off} 
\floatstyle{ruled}
\newfloat{listing}{tb}{lst}{}
\floatname{listing}{Listing}
\title{Scaling Laws Do Not Scale}
\author {
Fernando Diaz,
Michael Madaio
}
\affiliations {
Google Research\\
diazf@acm.org, madaiom@google.com
}

\usepackage{bibentry}

\begin{document}

\maketitle

\begin{abstract}
Recent work has advocated for training AI models on ever-larger datasets, arguing that as the size of a dataset increases, the performance of a model trained on that dataset will correspondingly increase (referred to as “scaling laws”). In this paper, we draw on literature from the social sciences and machine learning to critically interrogate these claims. We argue that this scaling law relationship depends on metrics used to measure performance that may not correspond with how different groups of people perceive the quality of models' output. As the size of datasets used to train large AI models grows and AI systems impact ever larger groups of people, the number of distinct communities represented in training or evaluation datasets grows. It is thus even more likely that communities represented in datasets may have values or preferences not reflected in (or at odds with) the metrics used to evaluate model performance in scaling laws. Different communities may also have values in tension with each other, leading to difficult, potentially irreconcilable choices about metrics used for model evaluations---threatening the validity of claims that model performance is improving at scale. We end the paper with implications for AI development: that the motivation for scraping ever-larger datasets may be based on fundamentally flawed assumptions about model performance. That is, models may not, in fact, continue to improve as the datasets get larger---at least not for all people or communities impacted by those models. We suggest opportunities for the field to rethink norms and values in AI development, resisting claims for universality of large models, fostering more local, small-scale designs, and other ways to resist the impetus towards scale in AI.\looseness=-1
\end{abstract}

\section{Introduction}
\label{sec:introduction}
In the context of multilingual language modeling,  the \textit{curse of multilinguality} refers to the degradation of model performance across languages as the diversity and size of the training data grows \cite{curse:conneau-etal-2020-unsupervised,curse:chang-etal-2023-multilinguality}.  These results suggest that bounded model capacity forces competition amongst languages for model parameters, leading to degradation of performance for \textit{all} languages.  In situations where languages are not uniformly represented, low resource languages tend to suffer more \cite{curse:wu-dredze-2020-languages}.  Similar results have also been observed in the context of mixed-modal models \cite{curse:aghajanyan-multimodal}.
We argue that the curse of multilinguality phenomena extends beyond language and modalities to include the values of different communities or sub-groups of people.  For large pre-trained models, model developers and evaluators often do not know about the diverse space of sub-groups, nor how to measure a model's performance for each community, making any claims of improved model performance limited to the values of an over-represented, high-resource sub-group.

We ground our analysis in the development of \textit{scaling laws} for AI: the relationship between machine learning model performance (as measured by some evaluation metric) and model design variables (e.g., dataset size, number of model parameters, compute) \cite{kaplan:scaling-laws}.
Existing research suggests that this relationship is  superlinear for deep learning models, and it has been used to justify the collection of ever-larger datasets  used to train large language models \cite{rae:scaling-llms,chowdhery:palm}.

However, the validity of these scaling laws relies on the extent to which the metrics used in scaling law analyses actually reflect the quality of a given model's performance for all groups of people impacted by large AI models deployed at scale.
In the context of AI systems that are used by or impact people, an evaluation metric (the dependent variable underlying many scaling laws) is intended to measure the performance or quality of the system \textit{for those people}.
The quantity computed by an evaluation metric is shaped both by its underlying mathematical assumptions and by the sampling procedures used to collect data to compute it. On one hand, the \textit{mathematical form} of an evaluation metric ideally reflects what a system designer thinks is important to users or others impacted by the system, encoding assumptions about users' beliefs, values, and behavior. On the other hand, the \textit{evaluation data} used to compute the metric is meant to capture some sense of the quality of that model by sampling a particular population of interest. In combination, the mathematical form of a metric and the associated data used to compute it estimate the performance of a model for a specific population, as is especially salient in emerging learned evaluation metrics \cite{rei-etal-2020-comet,stanojevic-simaan-2014-beer}.

Therefore, when designing for a sufficiently diverse population of people, measuring performance with a single, universal metric may lead to misplaced trust in models' quality.
Populations are composed of \textit{subpopulations}:  members of particular identity groups, including demographic groups (e.g.,  gender, racial or ethnic groups, religion, caste, national identity, disability status), as well as other cultural or sociopolitical groups.
We argue that as the size of the population being designed for increases, the number of subpopulations present in any evaluation  is likely to increase.
Samples used to evaluate model performance will  
capture some set of communities in the social context in which the data was collected. 

Increasing the number of impacted people
and the evaluation dataset size may thus increase the number of subpopulations present in the data and correspondingly increase the range of values that should be considered, potentially leading to values tensions \cite{miller2007value,ghassemi2022machine,birhane:values-in-ml,durmus2023towards,sorensen2023value,varshney2023decolonial}.  Beyond the well-known challenges of developing AI systems to be used by diverse subpopulations---such as evaluating algorithmic unfairness in model performance with respect to a fixed evaluation metric \cite{hardt2016equality}---we argue that diverse subpopulations are likely to have different and potentially conflicting notions of `good' \cite{green2019good} as instantiated in both the mathematical form and data behind an evaluation metric, which machine learning evaluations using single metrics often assume to be universal.
While algorithmic unfairness surfaces the systematic variation of the true relationship between features and a single universal metric across subpopulations, we contend that the metrics themselves may vary across subpopulations.

The current lack of attention to the subpopulations and communities represented in scaling law evaluation datasets
poses major challenges for proponents of AI scaling laws---and presents potential risks for all those impacted by the deployment of large models. Despite claims that a larger training dataset
will lead to improved model performance, when such models are deployed at scale, the larger numbers of people who are impacted by such systems or included in the evaluation dataset may lead to breakdowns in model performance for different communities \cite{bender2021dangers}. Different communities of users or impacted stakeholders may necessitate evaluating with different metrics (due to different behaviors or values), each of which may be in conflict with each other, or in conflict with commonly used evaluation metrics.

We draw on empirical results from machine learning and scholarship from the social sciences to interrogate the validity of claims made about scaling laws for training dataset size and model performance when considering their use in models deployed at scale, on global populations whose values may not be reflected in current performance metrics, or which may be in irreconcilable tension with values of other communities. These observations suggest that scaling laws, in the (perhaps misguided) pursuit of general principles, can obscure systematic under-performance for subpopulations and communities of people impacted by these models.

We thus propose that proponents of scaling laws re-consider their claims to universality of model performance across increasingly global populations of users of such large models. This may involve rethinking choices about evaluation metrics and the culturally-situated nature of evaluation datasets used to compute them; tempering claims for general applicability of scaling laws' evaluations of model performance; and making intentional choices about which communities are involved in evaluating models and how their values may be reflected in evaluation choices, including investigating how to resolve tensions between multiple communities' values when evaluating large models deployed at global scales. This paper contributes to a growing body of scholarship that explores the impacts of large models on global populations.

\section{Scaling Laws}
\label{sec:scaling-laws}
In order to scrutinize scaling laws from the perspective of evaluation metrics, we first review the relevant concepts from AI evaluation and scaling law literature. We will introduce some concepts from measurement modeling  \cite{hand:measurement-theory,jacobs:measurement-and-fairness} and introduce some notation that we will refer to throughout our analysis of scaling laws.

\subsection{Performance Metrics are Imperfect Proxies for Model Quality}
\label{sec:scaling:metrics}
Evaluation of AI systems involves computing and comparing quantitative measures of performance of a system on a task.  In offline settings, including laboratory or benchmark experiments, researchers use evaluation metrics based on data labeled through dedicated annotators.
In online settings, including deployed AI systems in production environments, organizations use evaluation metrics based on logged behavior data.
In both settings, evaluation metrics play a critical role in guiding high-level research and model development decisions as well as more granular parameter optimization \cite{Cohen:book,kohavi:book,joachims:ranking-svm,grotov:online-ltr-tutorial}. In particular, offline metrics are often used---implictly or not---as proxies or predictors of online metrics \cite{zheng:eval-book,suresh2021framework,Rudin:2014aa}.  In line with modern AI paradigms, we focus on the evaluation of models trained on a set of data.  We represent a model trained  on a dataset $\trainset$ as $\trainedmodel$. For the purpose of our analysis, we are interested in the relationship between data and model performance and therefore will not specify the input or output space of $\trainedmodel$, nor do we care about the specific functional form or design of $\trainedmodel$.  We use  $\modelspace$ to refer to the space of all trained models, of which $\trainedmodel$ is one member. For clarity, we will sometimes refer to a model as $\model$, even though it is always the outcome of a training procedure and data.\looseness=-1

Generic metrics like accuracy or squared error are often adopted in machine learning contexts to evaluate the quality of the model output. These metrics are simple, well-understood, and usually amenable to direct optimization. For many AI systems, however, generic metrics do not capture how \textit{people} use the system output.  For instance, for a language model used for question-answering, how reflective is the measure of `accuracy of predicting the next word' of the underlying \textit{user goal} of `understanding the answer to a question'? For a recommender system, how reflective is the measure of `accuracy of predicting a rating' of the underlying \textit{user goal} of `discovery of new, relevant content'?  There is subtlety in how the system output is used by people, which is lost when performance is measured with generic metrics. This is why, in applied contexts, domain-specific metrics are usually developed.  Indeed, a variety of areas of research, including natural language processing (e.g., BLEU \cite{reiter2018structured,post2018call}, ROUGE \cite{ganesan2018rouge}), search and recommendation \cite{diaz:neurips-2020-tutorial}, and more \cite{raji2021ai}, have adopted families of metrics informed by technology use \cite[for a critique of NLP metrics, see][]{subramonian2023takes}.

Even within a given domain, the quality of a model's output may be tightly coupled with a user's specific task; for example, the quality of a predictive typing application may be related to how useful users find it in effectively completing a writing task---but this quality may differ greatly between different use cases for the same task, such as (for instance) informal messages and creative writing tasks compared to professional communication or scientific writing tasks. For AI systems, we are thus interested in evaluation metrics that measure---explicitly or not---the quality of a system's output (e.g., predictions, decisions, recommendations) for a given \textit{population of people} for a given \textit{task}.

Drawing on the measurement modeling literature \cite{jacobs:measurement-and-fairness,hand:measurement-theory}, we refer to the unobservable  domain- and task-specific notion of model output quality as a \textit{construct}, or, the true performance of a system when used by a specific population for a specific task. While in many cases, a construct is \textit{completely} unobservable, in others it may merely be very expensive to collect or more accurately estimate (e.g., downstream revenue effects; user satisfaction).  We define $\trueperformancebase$ as the \textit{latent} scalar value associated with the quality of $\trainedmodel$ for a population $\people$.  This population could be compact and well-defined (e.g., `coworkers in a specific academic department') or vague and general (e.g., `any person with access to a computer, regardless of location, race, ethnicity, age, religion, gender identity, sexual orientation, disability, or economic status'). Although it may be difficult or impossible to observe the latent construct---both because $\construct$ and $\people$ are unobserved---we can ``operationalize'' it using an observable \textit{evaluation metric} as a proxy \cite{jacobs:measurement-and-fairness,subramonian2023takes}.  For example, we can evaluate the quality of a text prediction model's output in terms of its usefulness for a given writing task by, for instance, measuring composition time, number of edits, the number of accepted text prediction suggestions, or using delayed user feedback on the quality of the text predictions \cite{robertson2021can}---but all of these are only ever proxies (some better than others) for the latent construct of interest, $\construct$.

To make this clear, let $\performancebase$ be the evaluation metric for a sample of user-associated data $\samplepeople\sim\people$ (or, a sample $\samplepeople$ from a population $\people$). User-associated data, often referred to as an `evaluation dataset,' may consist of individual data (i.e., each data point or instance is a person) or derived data (e.g., each data point is a text document written by a person).
So,  an evaluation metric approximates both the functional form of $\construct$ as well as the associated target population $\people$. In the next section, we discuss the ways that populations might impact the validity of evaluation metrics.
Let $\metrics$ be the set of \textit{all} evaluation metrics, (including better and worse proxies for the latent construct), that can be used for measuring system performance.  
From measurement modeling, the validity of an evaluation metric $\metric$ is based on the extent to which it captures the salient aspects of the latent construct $\construct$ it is designed to operationalize \cite{jacobs:measurement-and-fairness}.  The better the metric captures the construct, the better able that it is to assess the performance of a given model.  Thus, the relationship between what we may want to measure about AI systems (e.g., $\construct(\people,\trainedmodel)$) and the way we measure that (e.g., $\metric(\samplepeople,\trainedmodel)$) may not be straightforward.

\subsection{Scaling Laws: Learning Curves and Power Laws}
Scaling law analyses use ``learning curves'' to represent the relationship between system performance and training dataset size (i.e., the horizontal axis is a sequence of training datasets ordered by increasing size  $\trainsetsizei{i}<\trainsetsizei{i+1}$).  In general, proponents of AI scaling laws 
argue that $\construct(\people,\model(\trainset_i))<\construct(\people,\model(\trainset_{i+1}))$---i.e., that models trained on larger training datasets will perform better according to a given performance metric. However, such claims are often made  about the unobservable construct of a model's performance ($\construct$), but they are \textit{based on} observations from proxy metrics ($\metric$) instead of the construct ($\construct$) and population samples ($\samplepeople$) instead of the true target distribution ($\people$), as constructs and the full population may be difficult or impossible to directly observe \cite{ganguli:predictability,rish:scaling}.

The precise relationship between $\trainsetsizei{i}$ and $\metric(\samplepeople,\model(\trainset_i))$ often follows a power law; i.e., as long as it is not bottlenecked by its capacity or access to compute resources, a model's performance improves superlinearly as a function of training dataset size
\cite{rosenfeld:scaling-laws,kaplan:scaling-laws}.   Although initial results demonstrated scaling laws for natural language processing tasks, similar laws have been developed for multimodal and reinforcement learning models \cite{cherti:language-image-scaling,hilton:rl-scaling-laws}.
The evidence of scaling laws has motivated their usage in informing model design.  They have been used as motivation for scraping ever-larger training datasets \cite{dodge2021documenting},  to developing quantization methods \cite{dettmers:four-bits-scaling},  to extrapolating performance from smaller datasets \cite{ivgi:scaling-microscope}, and to determining data minimization policies \cite{shanmugam:dm}. Despite their popularity, however, insufficient attention to the precarious relationship between latent constructs of model quality and their operationalization in performance metrics \textit{for particular communities included in evaluation datasets} (or who are otherwise impacted by a model) poses serious questions to the validity of relying on scaling laws when evaluating deployed models at scale.
\section{The Precarity of Metrics}
\label{metrics}

In order to understand precisely how scaling laws might be compromised,  we first need to review the various ways in which metrics are far from the `ground truth' they are often considered to reflect.
Through a  discussion and review of existing observations in the computer science and social science literature, we  demonstrate that evaluation metrics are inherently contestable  and precarious and that, at any one point, there may be multiple metrics and constructs in tension.
We draw on and extend recent work in the responsible AI and computational social science communities that similarly recognizes that, although  presented as  a reliable proxy for the construct, evaluation metrics are  often tenuous  \cite{friedler:fairness-impossibility,jacobs:measurement-and-fairness,thomas:reliance-on-metrics,wagner:measuring-algorithmically-infused-societies}.\looseness=-1

\subsection{Metric (In)compatibility}
\label{sec:metrics:compatibility}
As previously discussed, an evaluation metric can be seen as an approximation of the construct of interest.  A generic metric like accuracy  is often adopted because it can be used without modification.  But, for any given task, a construct
is more complicated than suggested by a generic metric. It might be impacted by a user's expectations, interface constraints, or normative social values, none of which are captured by a generic metric like accuracy. Because metrics are errorful proxies for constructs, any two metrics, even if they largely agree with the construct, may disagree with each other.  Moreover, because constructs and their operationalizations may be contested, two metrics may disagree because each is modeling a different (though related) underlying construct (e.g., two fairness metrics may be operationalizing different understandings of fairness \cite{chouldechova:fairness-impossibility,jacobs:measurement-and-fairness}).
Thus, adopting a single metric for evaluating a model---including for scaling law analyses---requires validating it with respect to the construct of interest, and how that relationship may be more or less stable for various use cases, social contexts, or even different ranges of the metric's value.  Moreover, understanding the relationship \textit{between} multiple metrics for different ranges of the metrics' values helps designers avoid situations where a metric becomes unreliable as model performance improves.
\looseness=-1

\subsection{Metric Nonstationarity}
\label{sec:metrics:nonstationarity}
The ability of a metric to capture a construct of interest may also change over time.
First, due to the complexity of sociotechnical systems made up of deployed AI systems used by or impacting people, metric development---like model development---is iterative, based on our understanding of how people and social groups engage with the technology \cite{neurips-2020-tutorial:beyond-accuracy}, as has been found in  web search evaluation \cite{carterette:evaluation-tutorial} and multidocument summarization evaluation
\cite{shapira-etal-2021-extending}.
At any point in time, a specific metric  captures the researcher's or practitioner's  best understanding of how to model the construct. But, because our understanding of users and the world is changing, this metric will also change over time \cite[cf.][]{subramonian2023takes}.
Second, as suggested by the possible heteroscedastic relationship between metrics and constructs, the `appropriate metric' at any point in time may depend on the set of systems being compared. 
Because model performance will change---ideally improve---over time, metrics will become stale and require replacement with other, more appropriate metrics \cite{voorhees:too-many-relevants}, due to the fallibility of proxy metrics with respect to a target metric or construct \cite{gao:scaling-overoptimization,skalse:reward-hacking}.
Third, model development often exists in a broader sociotechnical context, which itself might change over time.  Consider a metric such as `consumption time,' used in a number of media platforms to model relevance by measuring how long someone spends consuming media.  The higher the consumption time of a particular piece of media, the more relevant the item is to that user.  However, improvements to bandwidth and rendering times on devices can  result in substantive changes in the relationship between  the consumption time and relevance.  Or, acute shocks to the environment (e.g., unexpected exogenous events, such as disasters, or regulatory policy changes) can  change how people, technology, and their measured values behave and, as a result, relate to a construct of interest \cite[cf.][]{selbst2019fairness}. In some cases, the model itself may lead to behavioral or environmental changes that impact the usefulness of the metric.  For example, users of a system may, over time, adapt their behavior in ways that compromise assumptions in the metric (e.g., how people express queries on mobile search engines \cite{kamvar:mobile-search-1,kamvar:mobile-search-2}).  Or, so-called adversaries may attempt to manipulate the system by gaming the metric (e.g., clickbait \cite{chen:clickbait}).

\subsection{Metric Staging}

Multiple metrics may arise for pragmatic reasons.  In model development,  different metrics are used at different stages due to the costs involved in metric computation \cite{zheng:eval-book,suresh2021framework}.  In production systems, behavioral metrics based on real interactions with end users (e.g., user engagement metrics) may be more accurate reflections of a latent construct (e.g., due to face validity \cite{jacobs:measurement-and-fairness}), but more expensive because---in addition to potential costs of deployment or data collection---they incur a risk of harming the users, losing their trust, and having them abandon the technology altogether \cite{wang2023designing}.  On the other hand, in offline evaluation, metrics based on data labeled by annotators (e.g., precision, recall, simulations) are less accurate reflections of the construct than behavioral metrics.  That said,  they are extremely fast to compute, often reusable, and can be used in an isolated environment where unexpected or harmful outputs may be insulated from impacting users.\footnote{However, harmful outputs are unfortunately often not prevented from impacting annotators \cite[cf.][]{gray2019ghost}.} Unfortunately, offline evaluation requires an annotation budget and the development of guidelines for annotators, both of which can be prohibitive.
In practice, model development involves coordinating multiple metrics: using  offline evaluation and associated metrics for a large set of models before selecting a subset for production evaluation using a different set of metrics.\looseness=-1

\subsection{Metric Variation Across Subtasks}
Although an AI system intended for a specific task or domain is often evaluated using a single construct or metric, in reality, there are often a diversity of specialized subtasks or subdomains.  For example, both question answering systems  and search engines support a wide variety of possible user information needs \cite{broder:web-search-taxonomy,murdock:question-triage}.  In turn, each subtask or subdomain has a unique construct.  This may manifest from some property of the labels (e.g., contextualizing labeled data on the subtask or subdomain) or the formula for the metric itself (e.g., in the search context, reciprocal rank for navigational intents versus recall for literature review).  In some situations, subtask metrics may be compatible or `compatible enough', allowing for shared information when  evaluating a model \cite{zhou:task-similarity-theory}.  In other situations, two subtasks may be have quite different constructs in tension \cite{aribandi:ext5}.  While a single, universal metric may be one way to resolve inconsistency amongst subtask or subdomain constructs or metrics, in a reinforcement learning context, \citet{skalse:reward-hypothesis-is-false} demonstrate that jointly optimizing a single policy by reducing multiple rewards (i.e., metrics) to a single number is only possible in a limited set of situations.
However, despite this variation across tasks, the dominant paradigm used in scaling law analyses is to optimize and evaluate for a single performance metric. \looseness=-1

\subsection{Metric Power} 
\label{sec:metrics:organizations}
When metrics are used to guide research in academic communities, make decisions in industry, and evaluate performance of machine learning systems, they become social objects amongst researchers, funding agencies, engineers, product managers, and other individuals involved in the processes of machine learning research, development, deployment, and use.  As such, metrics used to evaluate AI systems are always subject to (and in turn, shape) the social dynamics of the sociocultural systems in which they are embedded.
When metrics are used to measure phenomena in social contexts, it is well-established that they not only enable one to understand a particular phenomena, but they also have the power \cite[cf.][]{beer2016metric} to change social actors' behaviors. In other words, metrics do not simply reflect the world, but they shape it, and are shaped by it  \cite[cf.][]{pinch1984social}. This phenomena, known as Goodhart's Law \cite{goodhart1984problems,campbell1979assessing,strathern1997improving}, has been identified across numerous fields, including education \cite{strathern1997improving,griesemer2020taking}, economics \cite{mugge2022economic}, and organizational studies \cite{gray2015measurement}. For instance, when schools use students' test scores as a metric to evaluate teacher quality, some teachers and administrators have responded by teaching to the test or in the worst case, altering students' test scores \cite{gabriel2010under}.
In the workplace more generally, companies have long adopted performance metrics of various sorts to measure their employees' productivity \cite{ranganathan2020numbers} and attempt to incentivize them to be more productive \cite{zelizer1996payments}---metrics that are increasingly based on fine-grained data from those employees, and which may shape employees' behaviors in other ways (e.g., employees using applications to move their cursor to simulate productivity) \cite{bernstein2017making}.
However, the specific ways that metrics impact people's behavior are themselves shaped 
by the norms, culture, and organizational dynamics of the context in which they are used. For instance, \citet{christin:metrics-at-work} found that two newsrooms in France and the United States that had access to similar web traffic data about their news stories differed greatly in how the metrics they derived from those data shaped their approach to journalistic decision-making, in ways impacted by their respective organizational, professional, and cultural contexts. 
In addition to the culturally specific ways that metrics may impact particular social worlds, metrics may take on a social life of their own. Within research communities, metrics can have a stickiness when entire research programs develop around them \cite{cohen:how-evaluation-guides-ai-research}.\looseness=-1

All of this suggests that the performance metrics used to evaluate AI systems may be inherently unstable or precarious in ways that raise serious questions for the validity and robustness of scaling laws for AI, which rely on metrics that are often divorced from any particular social context.

\section{Values Pluralism Threatens Scaling Laws at Scale}
\label{claims}
We now turn to critically interrogate scaling laws in light of the multiplicity of human values across the numerous social groups and communities impacted by models and represented in evaluation datasets.  While there are existing initiatives to identify `inverse scaling laws' for particular tasks \cite{wei:u-shaped-scaling}, our claim is that, when we consider training and evaluating with human data, scaling laws, as currently posed, are, at best, incomplete, and may be fundamentally flawed. Our claim is divided into four parts. First, that evaluation metrics reflect the composition of the evaluation dataset, which is shaped by the sampling approach used to collect that data; second, that the number of sub-groups within a given evaluation dataset grows with data size; third, those sub-groups can have incompatible values and preferences for appropriate evaluation metrics; and fourth, that the risk of that metric incompability grows with dataset size.\looseness=-1
\subsection{Sampling Approaches Shape who is Included in Evaluation Datasets}
\label{sec:claims:sampling}
As we mentioned earlier, metric-based evaluation  estimates the performance of a model when used by an \textit{intended} population of users.  Consistent with standard assumptions in machine learning, ideally both training and evaluation data are drawn from the same population using the same sampling distribution (although in practice this may not be the case).
If  $\people$ is the population, then $\samplingdist$ defines the sampling distribution from which we draw both the training set $\trainset$ and the evaluation set $\samplepeople$.  When describing an evaluation metric, we can describe it in terms of the sample size $|\samplepeople|$.  For instance, benchmarking or offline experiments have a fixed sample (and  sample size) determined by the collected, static evaluation data, where training and testing sets are often defined by by a percentage of instances (e.g., 80\% training; 20\% testing).  In deployed systems, the sample size varies with the number of users (and the training data size $|\trainset|$).

While we have talked about models evaluated when used by or impacting people for specific tasks, we have  avoided the question of \textit{which people} an AI system is designed for, used by, or impacts.  Given a current set of users (e.g., in a deployed system), we can answer this question narrowly by evaluating a model with respect to the existing set of users of a system or application that a given model is embedded in; in this case, we say that the sampling distribution has ``support'' constrained to the subpopulations reflected by the current set of users. Or, we can answer this aspirationally by evaluating with respect to some future population of users; in this case, we assume that the sampling distribution has support that bounds all subpopulations present in the complete population. For the purpose of our analysis, because of the aspirational nature of claims made about AI scaling laws \cite{kaplan:scaling-laws}, we assume that the support of the sampling distribution is complete. 
This implies that every possible user has a nonzero probability of occurring in the training or evaluation data.\looseness=-1

However, regardless of the sampling distribution, in any specific sample, we rarely (if ever) have reliable data for every possible user in every context. Nevertheless, the desire for such a dataset has led researchers to seek out larger sources of data on human populations \cite{hargittai2015bigger,yoo2015not, kitchin2014big,anderson2008end}.
Debates about whether and how different groups of people may be represented by or within large datasets have persisted for decades, across multiple fields \cite{chasalow2021representativeness, bergman2023representation}. Social scientists conducting public opinion surveys have long wrestled with what it means for their samples to be representative (and of whom they might be representative), including, for instance, examining representativeness of survey respondents from various demographics across geographic scales (e.g., cities, states, or national populations) \cite{de2016computational}. Many theories and methods in the social sciences grappled with the fundamental heterogeneity of human populations, as groups of people may vary based on gender, age, race/ethnicity, disability status, as well as behavior, physiology, language, religion, culture, and numerous other dimensions \cite{feczko2019heterogeneity}. These dimensions of difference may themselves be more or less stable, as people may claim membership in numerous communities at different points in their lives, as such identities become more or less salient (e.g., a person leaving or joining a religion, moving to a new city or country, joining or leaving the military, and numerous other examples). In other cases, the community itself may spring into existence in response to a particular political issue (e.g., DREAMers, NIMBYs, climate change or election deniers, and more) \cite{disalvo2009design}.\looseness=-1

Depending on one's question of interest, different approaches may be needed to understand how a sample captures variation across a larger population \cite{feczko2019heterogeneity}. That is, the means we use to capture these different dimensions or communities are often designed based on the goal: political surveys may be designed to capture population-level variation across politically salient demographic categories, such as gender, race/ethnicity, education level, and income, while marketing research surveys may be designed to specifically target consumer-relevant demographics and behaviors, such as family size, technology use, spending behaviors, and more \cite{de2016computational}. Although many datasets about people, such as opinion polls, are intentionally designed to answer specific research questions, the advent of large-scale data collection enabled by data traces on digital platforms has led computational social scientists and others to explore how such datasets may enable them to better understand people \cite{yoo2015not, kitchin2014big}. Despite claims for massive datasets (in the form of `big data') to usher in ``the end of theory'' \cite{anderson2008end} via datasets where, allegedly, ``\textit{n $=$ all}'' \cite{kitchin2014big}, subsequent research has demonstrated that large-scale platform datasets are always reflections of behaviors of particular groups of people (rather than being somehow inclusive of everyone) \cite{boyd2012critical}.\looseness=-1

As large datasets from social media platforms (e.g., Twitter, Facebook, Reddit, etc) are used to investigate research questions not only about platform use, but about social dynamics more generally, these platforms may suffer from a `model organism' problem \cite{tufekci2014big}, where claims are made about the world in general, based on data from one specific organism, such as mice or fruit flies, or one specific social media platform or datatset. In reality, there is a non-random selection of users into social media platforms \cite{hargittai2015bigger}; for example, Twitter is used by less than 20\% of the US population \cite{mitchell2013twitter}. An empirical analysis of who is left out of so-called ``big data'' from social media found that social media users tend to be more educated, higher-income, and more technologically savvy than non-users, with substantial differences in gender and race/ethnicity across different platforms. \cite{hargittai2020potential}. Similarly, massive datasets used to train large models, such as the Colossal Clean Crawled Corpus (C4), trained on a crawl of the web, or others such as LAION \cite{luccioni2021s,birhane2021multimodal}, are not representative of everyone on the planet, but contain particular snapshots of particular populations (and not others) \cite{bender2021dangers}. For instance, some communities may not be included at all in the C4 corpus or other large web-based corpora: low-literacy communities or those who rely on radio for information and communication \cite{khan2017role}; communities with low or no technology use---or those whose technology use is primarily on mobile devices, and does not involve producing web-based content legible to web crawlers \cite{okeleke2019mobile,aker2010mobile}; or communities who speak low-resource languages \cite{magueresse2020low}, among many others \cite{bender2021dangers}.\looseness=-1

As such, it is an open question as to who, precisely, is represented in massive datasets used to train and evaluate models for scaling laws \cite{bergman2023representation,bender2021dangers}. The composition of a dataset depends on the sampling approach used to create it---whether that is a random population sample, a sample stratified by some dimension(s) (e.g., randomized within different state populations), or a convenience or platform sample (i.e., a sample shaped by the nature of the platform used to collect the data, such as Twitter, Reddit, or the web). For all of these approaches, we cannot say with certainty precisely how many communities or sub-groups are reflected in a given dataset without access to information about its sampling or data collection methodology. Even then, because the definition of a given community and an individual's membership in it may be fluid over time or potentially overlapping or intersectional \cite{crenshaw1990mapping,collins2019intersectionality,ovalle2023factoring}, the number of communities represented in a dataset may depend on the research questions and methods used to investigate that question.

\subsection{The Number of Sub-Groups in Evaluation Datasets Grows with Dataset Size}
\label{sec:claims:subgroup-growth}
Since evaluation metrics are influenced by the sampled population $\samplepeople$, which itself may be non-representative for a variety of reasons, often related to the sampling approach, we now discuss how, as we increase the sample size, the number of sub-groups present will also increase. Consider a theoretical sampling distribution that aims to be representative of the population $\people$.  In this case, the sequence of samples $\samplepeople_0,\samplepeople_1,\ldots$ will, although initially missing out on many sub-groups, converge toward covering a broad set of people, ostensibly representing all sub-groups in $\people$, which is consistent with the aspirational, though unrealistic, claims  for large models to benefit ``all of humanity'' or ``the human race'' \cite{altman2023planning}.  The growth in heterogeneity as a function of population has similarly been studied in political science contexts \cite{dahl:size-democracy}.
As mentioned, the sampling frames typically used to evaluate AI systems are \textit{not} intentionally collected to be representative of any particular sub-group or community---nor is it clear what it would even mean for a sampling frame to be representative of all sub-groups of people \cite{chasalow2021representativeness}.

This means that although smaller samples are likely to contain fewer sub-groups and a large enough sample may theoretically converge toward a broad, global set of sub-groups (with the caveat that some groups may not be known \textit{a priori}, or may only emerge due to exogenous causes after data has been collected), the rate at which the number of sub-groups are encountered is dependent on properties of the sampling distributions, and the broader sampling approach taken.  Given the tendency for data collection practices to be biased toward communities that are easier for the researchers collecting the data to access \cite{madaio2022assessing}, or datasets that are easier to scrape \cite{dodge2021documenting}, the rate of observing new sub-groups will be slower than a representative sample.  While stratified sampling by subpopulation may suggest a possible solution, there are several reasons to be suspicious of this approach.  Enumerating all possible relevant subpopulations present in the broad target population imagined for AI is a daunting task, in part because of the large number of subpopulations but also because of the fact that they can emerge and dissolve over time.\looseness=-1

More realistically, if sampling is done by `organic user growth,' as is typical in production settings, the sampling distribution itself is changing as sample size increases.  Consider deployed systems, where early adopters of the technology will not be representative of later users.  For example, user growth on social media platforms tends to occur non-uniformly within and across national boundaries \cite{poushter:social-media-developing-countries,perrin:social-media-trends,wilkinson:myspace-users-over-time}.  Assuming consistent adoption and growth, we can use a model of nested subpopulation support, $\support(\samplingdist_0)\subseteq \support(\samplingdist_1)\subseteq\ldots\subseteq\people$.  As suggested by social media adoption, the growth in support is structured:  some populations tend to adopt before others due to homophily.

These observations suggest that, regardless of the sampling strategy or the way we might represent a sequence of samples (e.g., the model of nested support), the number of unique sub-groups present will grow with sample size. Crucially for scaling laws, the \textit{nature} of that growth---and thus the particular composition of sub-groups and communities contained in the evaluation dataset---is heavily impacted by the sampling strategy used to collect the evaluation dataset.\looseness=-1

\subsection{Sub-Groups Can Have Incompatible Metrics}
\label{sec:claims:incompatibility}

Along with the substantial heterogeneity of populations comes heterogeneity of preferences and values. Although different sub-groups may have different relationships with a single evaluation metric 
(as in dis-aggregated evaluations of algorithmic fairness \cite{barocas2021designing,rolf:data-externalities,burnell:disaggregate-metrics}), we are particularly interested in different, incompatible metrics and constructs themselves \cite{green2021contestation,jacobs:measurement-and-fairness}, a subtle but important difference. For instance, it is well-established via decades of the World Values Survey that there are tensions in values across international populations \cite{haerpfer2020world}. In addition, there is large sub-national variation in public opinion; for example, in the US, public opinion differs greatly on topics such as support for gay marriage \cite{caughey2019public}, the New Deal \cite{caughey2018policy}, and the death penalty \cite{shirley2015hierarchical}, among others \cite{berman2020measuring}, in ways that are shaped by various cultural and political factors \cite{caughey2019public}.

In AI ethics, prior work has identified substantial differences in how various populations' values manifest in terms of preferences for AI systems. For instance, \citet{jobin2019global} analyzed AI ethics principles statements from nearly a hundred different institutions across various countries, finding that while there was convergence in high-level values such as fairness and transparency, there was high divergence between countries in the specific ways those values are operationalized in AI principles statements, the practices that enact those principles, and the mechanisms used to enforce them. In addition, \citet{awad2018moral} collected data on millions of people's preferences for which of two personas an autonomous vehicle should kill in a car accident, via an online tool used in 233 countries and territories. They found substantial cross-cultural variation in preferences \cite{awad2018moral}, and they attempted to explain that variation by drawing on various economic and cultural indicators, such as the World Values Survey \cite{inglehart2000world}, Gini coefficient scores \cite{dorfman1979formula}, and other cultural frameworks \cite{mcsweeney2002hofstede,hofstede2011dimensionalizing}. 
Relatedly, \citet{jakesch2022different} conducted a survey of how different groups prioritize ethical values in AI development, finding statistically significant differences in how members of different occupations and demographic groups prioritize values such as fairness, privacy, and transparency in particular AI deployment scenarios. Recent work has explored the relationship between different groups' responses to public opinion polls (e.g., Pew American Trends and the World Values Survey) and the output of large language models, finding that large models' output is more similar to the \textit{average} responses from survey respondents in the USA, Canada, and Australia, compared to respondents from other countries \cite{durmus2023towards} and within the US, language models' output reflects certain groups' opinion poll responses more often \cite{santurkar2023whose}.\looseness=-1

As one example of how cultural differences in values may impact AI development and evaluation, \citet{sambasivan:fairness-in-india} has identified how algorithmic (un)fairness in the Indian context operates along different axes than those identified in Western contexts. For instance, they found that algorithmic fairness in India entails different sets of sub-groups, frameworks, and methods, including how algorithmic harms are shaped by the forces of caste and religion, among others. They discuss how popular fairness measurements are informed by specific cultural and historical circumstances, such as approaches for measuring disparate impact or disparate treatment arising from US anti-discrimination law \cite{watkins2022four}, and by Western philosophical frameworks and approaches to justice more generally \cite{sambasivan:fairness-in-india}.
This suggests that large-scale datasets (which may contain numerous communities or sub-groups) may thus inadvertantly collapse meaningful differences in those sub-groups' preferences for how values in AI are operationalized---leading to what some have referred to as ``aggregation bias'' \cite{suresh2021framework}. There is substantial empirical evidence for how such aggregate approaches to evaluating models may hide disparities between (or within) subpopulations for a fixed evaluation metric. To uncover these disparities, ``dis-aggreggated evaluations'' \cite{barocas2021designing} of model performance are conducted by dis-aggregating a single performance metric across multiple groups. Such approaches have been the foundation of high-profile evaluations of machine learning failures, such as evaluations of how gender recognition systems are less accurate for women of color than others \cite{buolamwini2018gender}, or how speech recognition systems have higher word error rates for speakers of African-American Language \cite{koenecke2020racial}, among other examples \cite{obermeyer2019dissecting, de2019does, ngan2020face}.\looseness=-1

However, prior empirical work found that when AI product teams deployed models ``at scale'' (i.e., across numerous geographic and cultural contexts), ambiguity about precisely \textit{how} to dis-aggregate evaluations of model performance---in terms of which evaluation metric to use, or along which demographic dimensions to dis-aggregate---posed major obstacles to AI teams' ability to effectively conduct fairness evaluations at scale \cite{madaio2022assessing}.  Indeed, concerns with aggregation bias are only amplified when we move from fixed evaluation metrics to subpopulation-specific metrics, which may be in tension with each other. Further contributing to potential aggregation harms, during the data annotation process, prior work has found substantial disagreement between annotators from different demographic groups when determining what constitutes hate speech or toxic language \cite{sap2019risk, thiago2021fighting, davidson2017automated}. Meanwhile, other work suggests that it is crucial to understand the subjective identities of crowdworkers \cite{diaz2022crowdworksheets,denton2021whose}, developing methods for handling disagreement between groups of annotators in situations where there may not be a single ``ground truth'' in annotation labels \cite{davani2022dealing, gordon:jury-learning,gordon:disagreement-econvolution}.\looseness=-1

Scaling laws, which are  aggregate evaluations of models' performance across the entire evaluation dataset $\testset$, may similarly hide failures or inverse relationships amongst constructs and values when evaluated with different sub-groups contained within the evaluation data.

\subsection{Risk of Metric Incompatibility Grows with Data Size}
\label{sec:claims:risk}
Given that the number of sub-groups  within the evaluation dataset grows with the size of that data, and these groups may have incompatible values (i.e., constructs for model output quality) and relationships to performance metrics that operationalize those constructs, we turn to our final claim: that the risk of failures or harms of AI systems grows as  data size grows.  
We ground this claim in the design and evaluation of AI, but similar observations have been made in the political science literature relating the size of a polity and an increase in value friction \cite{dahl:size-democracy}.

Goodhart's Law suggests that using a proxy metric can lead to over-optimization and degradation of the actual performance on the construct, something confirmed in the AI alignment literature \cite{gao:scaling-overoptimization,skalse:reward-hacking}.  Similarly, using a single dominant metric can lead to over-optimization and degradation of the actual performance for other constructs and values of different communities, especially because they are more likely to be incompatible with the dominant construct.  Indeed, work in multi-task learning has demonstrated that, when   optimizing for one task, more data can degrade performance for other tasks \cite{aribandi:ext5}.    In addition, \citet{solaiman2021process} find evidence for a scaling law between model size and toxicity---that is, as model size increases, the models were \textit{more} likely to generate toxic language. Similarly, \citet{lin2021truthfulqa} found evidence for an inverse scaling law for model size and truthfulness in a question-answering (QA) task (i.e., models were less truthful the larger they were), and \citet{parrish2021bbq} found that larger models performed worse on the task of detecting biased language, using a bias benchmark dataset they developed for QA. This phenomena has also been shown as the training dataset size is increased, in addition to the model size.  When analyzing the LAION datasets for the presence of hateful content in images and alt-text, \citet{birhane2023hate} found that as the dataset size increased, the likelihood for models trained on those datasets to label images of Black people's faces as criminals also increased.\looseness=-1

Since the number of distinct sub-groups (and thus their respective latent constructs for model quality) represented in  evaluation data is likely to grow with dataset size, there is an increasing chance of a dramatically misaligned evaluation of model quality---leading to potential impacts or harms for communities whose values are not represented by the dominant performance metric used for model evaluation \cite{ovalle2023m,dev2021harms,felkner2022towards}.

Given that we might observe a disparity in `true performance' across populations in the evaluation data or, more generally, in the target population, we need to quantify the severity of this disparity.  The systematic under-performance and exclusion of values from some sub-groups in scaling law analyses raises issues of (un)fairness and justice.  While we emphasize that our claims are different from those in the existing algorithmic bias literature that evaluate a fixed metric for different populations \cite{barocas2021designing,buolamwini2018gender}---as we are interested here in values tensions that might result in different communities valuing different constructs entirely or different proxy measures to capture those constructs---we can still adopt methods from that community to measure disparity \cite{barocas-hardt-narayanan}.  From the perspective of robustness or Rawlsian fairness, we can look at the worst case true performance of a system on a sub-group in terms of its own values \cite{liang:fairness-without-demographics,kearns:fairness-gerrymandering,ghosh:worst-case-intersectional}.  Poor performance is likely to be amplified by the fact that the worst off sub-groups  are likely to be from groups that historically have not been considered, represented, or participated in AI development processes \cite{queerinai2023queer}.  Given the catastrophic deterioration of performance according to Goodhart's law, other notions of unfairness (in terms of incompatibility of values and the metrics used to operationalize those values) are also more likely to occur as more sub-groups manifest in evaluation data and model performance according to a target metric grows.\looseness=-1

\section{Discussion}
\subsection{Implications for existing approaches to scaling laws}
\label{sec:disc:implications}

Proponents of scaling laws for AI systems argue for the existence of a power law relationship between the size of a model (i.e., number of parameters, dataset size, compute) and its performance (along some metric). While this narrative has led to increased investment in collection of large datasets \cite{luccioni2021s,birhane2021multimodal} and in ever-larger models and compute power---along with supporting a narrative of progress akin to Moore's Law---recent work has demonstrated that scaling laws may not hold for particular tasks \cite{wei:u-shaped-scaling,mckenzie2022inverse,caballero:broken-scaling,lin2021truthfulqa,parrish2021bbq}.

However, while the current work on exploring the limitations of scaling laws (e.g., inverse scaling laws \cite{mckenzie2022inverse}) has largely kept the same parameters---the relationship between some aspect of model size and model performance---and just uses a different \textit{task} (e.g., generating truthful text, negation, bias detection, etc), we argue that there may be other relevant dimensions along which scaling laws may not hold. For instance, we argue that scaling laws should consider the size of the evaluation dataset (in addition to the size of the training dataset).
In addition, given our argument  that different communities represented within a dataset (or impacted by a particular system) may have fundamentally different values and metrics, what might it look like to evaluate scaling laws where the vertical axis, instead of decontextualized model performance metrics like accuracy (or F1 score, RMSE, ROUGE, etc), were instead chosen for particular use cases or system deployment contexts, or were chosen by particular impacted communities in participatory ways \cite{wagner:measuring-algorithmically-infused-societies,delgado2023participatory,suresh2022towards, dennler2023bound} to better reflect their values. 

Returning to our introductory example, although mitigations for the curse of multilinguality have been developed \cite{curse:pfeiffer-etal-2022-lifting,curse:blevins2024breaking}, they require a well-defined set of languages and metrics for each, both of which are missing for the multiplicity of sub-groups  we suggest.

Moreover, substantial work on model evaluations has shown that aggregate metrics of model performance may hide worse performance for particular sub-groups that can be observed when model performance is \textit{dis}-aggregated by some demographic categories \cite{buolamwini2018gender,koenecke2020racial,obermeyer2019dissecting,ngan2020face,barocas2021designing}. Analogously, the current paradigm of evaluating scaling laws on aggregations of performance metrics evaluated on a single training 
dataset is likely to hide similar divergences in values and preferences for metrics for sub-groups within an evaluation dataset. For instance, a recent paper proposed a benchmark for evaluating bias in QA, and evaluated it on several large language models; however, they caveat at the end of the paper that ``\textit{the data in BBQ is only designed to test biases associated with US English-speaking cultural contexts, and it should not be used as evidence that a model would still look unbiased for contexts from a different culture}'' \cite{parrish2021bbq}. What would it look like for evaluations of scaling laws to be dis-aggregated for datasets collected from (or ideally collected, curated, or annotated \textit{by}) different communities \cite{wagner:measuring-algorithmically-infused-societies}, be that linguistic communities, cultural communities, countries, or other sets of sub-groups? In other words, would an observed scaling law relationship for performance metric $\metric$ evaluated with dataset $\testset$ still hold if that dataset were collected from a different context or collected or annotated by a different community?\looseness=-1

\subsection{Broader questions for scaling laws}

While the previous section suggests some shorter-term means to investigate the limits of scaling laws, this work raises more fundamental questions for scaling laws for AI. For instance, while analyses like the ones we proposed may reveal broken, inverse, or other non-monotonically increasing functional forms for scaling laws across different communities within a given dataset (or who might use or be impacted by a particular system), what to do about that is a much thornier question. In many ways, tensions in values (sometimes referred to as values pluralism \cite{berlin1969four,crowder2002liberalism,van2009public}) are a fundamental challenge of political systems---including technology, as technologies enact politics \cite{winner1987whale,mulligan2019procurement,nissenbaum2011preemption}. Methods have been developed from participatory democracy \cite{polletta2012freedom}, deliberation theory \cite{fishkin2005experimenting}, and value-sensitive design \cite{miller2007value} (among other areas) to identify and resolve tensions in values. However, these methods have largely been designed for smaller scales: e.g., town halls, focus group discussions, and participatory design workshops \cite{muller2012participatory}. As such, it is not clear how such approaches may be able to address value tensions at the scale of modern AI systems \cite{sloane2020participation,hanna:against-scale,delgado2023participatory,young2024participation}. This may be an argument for drawing on related work from anthropology that argues for ``non-scalability'' as a desired quality of social systems that resist the impetus to scale \cite{sharma2023post}, or that rethink the nature of systems' design as they move from context to context \cite{tsing2012nonscalabilitythe}, rather than relying on the supposed portability of sociotechnical systems \cite{selbst2019fairness}.\looseness=-1

Recent work on AI ``alignment'' has attempted to develop approaches to aligning AI with human values \cite{askell2021general,bai2022training,ganguli2022red,solaiman2021process}. \citet{gabriel:alignment} discusses how values pluralism may impact the goal of aligning AI systems with ``human values’’ (in all of their multiplicity and tensions), and he discusses tradeoffs in several potential approaches to resolving those tensions. Some currently adopted approaches include red-teaming \cite{ganguli2022red}, reinforcement learning from human feedback (RLHF) \cite{bai2022training} or creating ``values-targeted datasets'' \cite{solaiman2021process}.   However, in recent work attempting to incorporate values into training large models, people involved in red-teaming and RLHF appear to largely be US-based and may not be representative even of the US population \cite{ganguli2022red,bai2022training}. 
Meanwhile, in this example, the ``values-targeted datasets'' were created by the researchers, who acknowledge this limitation, writing that ``\textit{creating many values-targeted datasets to reflect the cultures of the many peoples impacted by language models is a difficult feat}'' \cite{solaiman2021process}.\looseness=-1

This is not, however, simply a limitation of their work, but a more fundamental challenge for values tensions in scaling laws---what might it mean to not only create values-targeted evaluation datasets from different communities or cultural contexts, but to resolve potentially irreconcilable differences in values between such communities, at scale? Is such a goal itself misguided?
While we proposed  a simple worst-case approach to quantify the robustness of a model across subpopulations, the reality of dealing with multiple, conflicting values is more complex.
The evidence from \citet{birhane2023hate} demonstrates that values of emergent subpopulations can be toxic, suggesting that simply looking at  subpopulations without context risks buttressing toxic behavior.\looseness=-1

Although we do not offer answers to the questions we have proposed in this paper, we suggest that, in part, what is needed are interdisciplinary, mixed-methods approaches to theoretically and empirically investigate the questions we raise. There are existing theories and methods from numerous fields---across the social sciences and computer science---that have been developed to explore questions related to whether and in what ways various communities or sub-groups are represented by data \cite{yoo2015not} as well as identifying and resolving values tensions among communities \cite{miller2007value}. For instance, various methods have been developed for identifying sub-groups within large datasets about people, in political science (e.g., for public opinion polling) \cite{feczko2019heterogeneity}, demography \cite{page2014diversity}, healthcare \cite{wick2022identifying}, genetics \cite{patterson2006population}, and more. 
However, work from the social sciences suggests that some communities may be hidden in ways not legible to data available for computational community detection (e.g., injection drug users; sex workers) \cite{salganik2004sampling}, requiring approaches such as qualitative or ethnographic research to identify such communities---which may be difficult and prohibitively expensive in current practice at the scales implicated by AI scaling laws, and moreover, many vulnerable or marginalized communities may not want to have data collected at all, as it may put them at risk \cite{muller2022forgetting,queerinai2023queer,dennler2023bound}.

In addition, recent work demonstrating the lack of relevance of Western AI fairness frameworks for other cultural contexts, such as India, has drawn on approaches from both qualitative \cite{sambasivan:fairness-in-india} and quantitative research paradigms, including natural language processing \cite{bhatt2022cultural}. We argue that interdisciplinary, mixed-methods approaches such as these, involving deep partnerships with community members or community organizations, such as participatory or community-based research methods \cite{delgado2023participatory,wagner:measuring-algorithmically-infused-societies,sambasivan:fairness-in-india,harrington2019deconstructing,muller2012participatory,dennler2023bound} may be one way to empirically investigate our claims and grapple with the inherent tensions and limitations of scaling laws for AI.

\subsection{Conclusion}

Analyses of AI scaling laws paint an impressive picture of progress in AI driven by increasing scales of data, model size, and compute power. We argue that these results are not the \textit{whole} picture. We draw together work from computer science and the social sciences to identify the ways that metrics used to evaluate
AI systems may be unstable or precarious; how the increasing scales of data used to train AI systems entail increasing numbers of sub-groups or communities of people; and how those groups' values and preferences for AI systems and the metrics used to evaluate them may be incompatible or fundamentally at odds with each other. We close by discussing how these insights pose fundamental challenges to the paradigm of AI scaling laws, and we raise open questions and opportunities for the field to investigate whether, when, and for whom scaling laws may (or may not) hold. We suggest opportunities for interdisciplinary, participatory, and community-based research to better understand which sub-groups or communities may be represented in a given dataset (or impacted by a particular model); which evaluation metrics might best reflect their values; and how to conduct such evaluations or resolve tensions in those values. The results of this work may suggest that progress in large AI models is not quite as straightforward as it first appears. In other words, one might ask: progress for whom?\looseness=-1

\bibliography{references}

\end{document}